\documentclass[preprint,12pt,authoryear]{elsarticle}

\let\today\relax
\makeatletter
\def\ps@pprintTitle{%
    \let\@oddhead\@empty
    \let\@evenhead\@empty
    \def\@oddfoot{\footnotesize\itshape
         {Preprint available on arXiv.} \hfill\today}%
    \let\@evenfoot\@oddfoot
    }
\makeatother

\usepackage[utf8]{inputenc}
\usepackage{verbatim}
\usepackage{amsfonts}
\usepackage{amssymb,amsmath,amsthm}
\usepackage{array, booktabs, makecell}
\usepackage{xcolor}
\usepackage{graphicx}
\usepackage{url}
\usepackage{multicol}
\usepackage{multirow}
\usepackage{subcaption}
\usepackage{natbib}
\usepackage{pifont}
\newcommand{\cmark}{\ding{51}} 
\newcommand{\xmark}{\ding{55}} 
\usepackage{subfiles}
\usepackage{soul}
\usepackage{threeparttable}
\usepackage[english]{babel}
\journal{International Journal of Forecasting}

\begin{document}

\begin{frontmatter}


\title{Hierarchical Neural Additive Models for Interpretable Demand Forecasts}

\author[a,*]{Leif Feddersen}
\ead{feddersen@bwl.uni-kiel.de}
\author[a]{Catherine Cleophas}
\ead{cleophas@bwl.uni-kiel.de}

\affiliation[a]{organization={Institute for Business, Christian-Albrechts-University Kiel},
                country={Germany}}
\affiliation[*]{Corresponding author}

\begin{abstract}
Demand forecasts are the crucial basis for numerous business decisions, ranging from inventory management to strategic facility planning. While machine learning (ML) approaches offer accuracy gains, their interpretability and acceptance are notoriously lacking.
Addressing this dilemma, we introduce Hierarchical Neural Additive Models for time series (HNAM). HNAM expands upon Neural Additive Models (NAM) by introducing a time-series specific additive model with a level and interacting covariate components.

Covariate interactions are only allowed according to a user-specified interaction hierarchy. For example, weekday effects may be estimated independently of other covariates, whereas a holiday effect may depend on the weekday and an additional promotion may depend on both former covariates that are lower in the interaction hierarchy.

Thereby, HNAM yields an intuitive forecasting interface in which analysts can observe the contribution for each known covariate.
We evaluate the proposed approach and benchmark its performance against other
state-of-the-art machine learning and statistical models extensively on real-world retail data. The results reveal that HNAM offers
competitive prediction performance whilst providing
plausible explanations.

\end{abstract}


\begin{highlights}

\item We introduce Hierarchical Neural Additive Models (HNAM) for time series forecasting. These models provide interpretability by clearly attributing how covariates affect forecasting outcomes while maintaining competitive predictive accuracy compared to advanced black-box models like Temporal Fusion Transformers (TFT) and clearly outperforming statistical methods.
\item HNAM's inherently explainable architecture, which composes forecasts from individual covariate effects according to a user-specified interaction hierarchy, has the potential to improve human-computer interaction in operational business planning by allowing practitioners to integrate their expertise and trust into the forecasting process.
\end{highlights}


\end{frontmatter}



\section{Introduction}
\label{sec:Introduction}

Demand forecasting as a problem of time series forecasting is at the heart of business planning activities \citep{mentzer2004sales} and, therefore, frequently subject to managerial oversight and intervention \citep{fildes2009effective}.
Contributions on judgmental interventions show that their overall impact on forecast accuracy is questionable while certainly costly \citep{fildes2021stability}, biased and inefficient \citep{fildes2009effective}, and highly dependent on the adjuster's expertise \citep{sanders1992need,edmundson1988use}.

One motivation for adjustments can be misaligned incentives across the organization \citep{mello2009impact}.
However, a general aversion to algorithmic advice \citep{dietvorst2015algorithm} is further thought to contribute to counterproductive human-computer interactions in forecasting tasks \citep{prahl2017understanding}.
One potential driver for algorithm aversion is that human decision-makers perceive automated algorithms as incapable of capturing the true demand dynamics \citep{lehmann2022risk}, or conversely
that opaque models prohibit model understanding, thus hindering trust \citep{adadi2018peeking}.

Classical, less opaque time series models such as (Triple) Exponential Smoothing, ARIMA, and Theta have long been viewed as superior to machine learning (ML) based forecasting approaches \citep{makridakis2018statistical}. Yet, recent results from forecasting competitions point to the superiority of ML methods for forecasting, particularly when time series are rich in covariates \citep{makridakis2022m5}.
Hence, while the increasing availability of high-dimensional data calls for complex ML models, their opaqueness could ultimately prevent organizations from realizing their potential due to lack of acceptance.

To address this dilemma for time series forecasting, we introduce a novel approach we term Hierarchical Neural Additive Models (HNAM).
HNAM combines the quasi-unlimited expressivity of neural networks and the interpretable form of additive models. 
Its forecasts are additively composed of a level as well as covariate effects that account for interactions across variables over time.
To that end, HNAM lets a neural network estimate the coefficients of a linear model where the model's available information set expands for each covariate according to a to a user-specified interaction hierarchy.
For example, given the hierarchy \emph{Weekday, Promotion, Holiday},
the estimated coefficient for the weekday effect would only depend on past observations, product attributes, and the weekday itself. The estimated promotion coefficient, one level up the hierarchy, would crucially account
for a weekday-dependent magnitude. 
Finally, the effect of a holiday depends on its weekday as well as scheduled promotions at the same time.
The obtained coefficients are finally multiplied by the feature values.
These feature values are one-hot encodings for categorical values whereas real values are preferably expressed as relative changes to a baseline (e.g. price deviation from the baseline) to foster sparsity and concise explanations.

Thereby, coefficients vary for each forecast and each horizon; HNAM estimates them via a black-box neural network able to model interactions and non-linearities. Yet, coefficients ultimately interact with the input features, constraining the model's expressivity and thereby its potential for overfitting.

Section \ref{sec:Motivation} motivates the HNAM architecture by discussing related literature on judgemental forecasting, algorithm aversion, and the impact of a decomposed transparent display of time series on acceptance and task outcomes. This section also introduces the existing models that have inspired our proposed architecture.

Section \ref{sec:Architecture} details the technical aspects and implementation details of HNAM. Section \ref{sec:datasets} introduces the three retail demand datasets and outlines the preprocessing used to benchmark HNAM against prevalent machine learning and statistical models. Subsequently, Section \ref{sec:results} presents the results of these benchmarking experiments and illustrates exemplary interpretability outputs.
We derive conclusions from these results in Section \ref{sec:conclusions}.
\section{Motivation and Methodological Background}
\label{sec:Motivation}

To motivate the proposed architecture, this section provides a brief overview on algorithm aversion and links it to judgmental forecasting. This motivation supports our argument that the interpretability output of HNAM could improve managerial interaction with its forecasts. Additionally, this section also provides a methodological background and outlines related approaches, namely Prophet, Neural Additive Models (NAMs) and Neural Additive Time Series Models (NATMs).

\subsection{Algorithm Aversion and Interpretability}
\label{subsec:algorithm aversion}

Research on the efficacy of algorithmic decision-making and the reluctance of human experts to accept algorithmic advice is extensive. It traces back to work by \citet{meehl1954} on medical predictions and by \citet{dawes1979the}, who showcase the superior performance of even basic linear models compared to human predictions. Despite these findings, practitioners frequently hesitate to rely on algorithmic advice. This reluctance is coined algorithm aversion in \citet{dietvorst2015algorithm}.
\citet{burton2020systematic} offer a comprehensive literature review and identify five major factors for algorithm aversion. In the remainder of this section, we summarize these factors and argue how they relate to our proposed architecture. We conclude that a lack of compatibility in how managers and algorithms approach the forecasting task seems to be at the core of algorithm aversion.

\paragraph{False Expectations} Decision-makers' trust in algorithmic advice is shaped by their own experiences and external reports, leading to different perceptions compared to human-generated advice. The desire for social interaction with the advice source, particularly among confident experts, can make algorithmic assistance seem unnecessary. People view human errors as random and correctable, while algorithmic errors are seen as systematic. Even if initially trusting an algorithm, unavoidable errors may lead decision-makers to preferentially rely on human decision-making, even despite recognizing the algorithm's superior performance. This aspect is largely independent of the model, except for the obvious goal of avoiding erroneous algorithms.

\paragraph{Lack of Perceived Control} In high-stakes situations, decision-makers want influence, independence, and the option to overrule or engage with algorithmic decisions. This necessitates the algorithm to be "behaviorally packaged" \citep[p.5]{burton2020systematic} to impart meaningfulness. However, the algorithm's underlying architecture determines the degree to which it can be presented meaningfully. In this contribution, we target this by designing HNAM to create composed forecasts. Instead of adjusting  outputs from an opaque architecture, users can exert control on the level of individual covariate effects.

\paragraph{Cognitive Compatibility} While human decision-making processes tend to be marked by instincts, heuristics, and biases, algorithms rely on deterministic, data-driven logic. Integrating the two calls for bridging these very different systems. Algorithmic interpretability could offer a solution, but at present, it only applies to simpler approaches and not to cutting-edge machine learning methods. The research presented here relies on the idea that managers tend to think about forecasts by adding and subtracting the expected impact of ex-ante known events to a baseline. Therefore, we design HNAM to feature an additive output layer - thereby matching the world model of the decision-maker. The preceding layers' exact logic are still opaque and non-traceable. However, we consider this aspect to mirror human experience insofar as humans come up with simple explanations of the world while the neurological processes that underlie our conscious thinking are phenomenologically in the dark \citep{nisbett1977telling}.

\paragraph{Lack of Incentives} Incorporating algorithmic advice in human decision-making can require substantial work which requires appropriate incentives. Economic rewards that encourage high-variance decision-making can deter the application of supposedly lower-variance algorithmic advice \citep{dietvorst2020people}. Social and organizational incentives may not align with the optimization function of the algorithm \citep{mello2009impact}. This aspect is largely model-independent and therefore not within the scope of this paper. However, future research, particularly focusing on behavioural experiments, could and should pick up on this aspect.

\paragraph{Divergent Rationalities} Humans and algorithms draw different subsets of information from their surroundings, process them in unique ways, and may have divergent objectives. In fact, 'broken-leg' cues (i.e., highly pertinent information that a human would know to incorporate into their judgment, but which a statistical model based on historical data would miss) are a common and sensible adjustment motive \citep{meehl1954}. HNAM's composed forecasts enable a more precise injection of extra-model knowledge by clearly differentiating considered covariates and their effect sizes.

\subsection{Judgmental Forecasting and Adjustments}
\label{subsec:judgmentalforecastsdecomp}
Unfortunately, behavioral research on improved acceptance and task outcomes of additively composed forecasts is scarce.
A related case study by \citet{vossing2022designing} finds that a more transparent forecasting interface, which lists the considered covariates for a forecast, results in more favourable adjustments.
The authors conclude that their transparent user interface allows managers to \emph{calibrate} their interventions more precisely and productively. That is, they become more attuned to the circumstances under which their extra-model information (or intuition) can, in fact, improve the forecast whilst abstaining from adjustments otherwise.

Further relevant studies on judgmental forecasts suggest that a decomposed presentation and processing of time series aids humans in creating their own forecasts:
Both \citet{lawrence1985examination} and \citet{carbone1985accuracy}, conduct a comparison of judgmental and statistical forecasts with M1 Competition data \citep{makridakis1982accuracy}. The former study finds that judgmental extrapolation matches the effectiveness of statistical extrapolation, while the latter reveals a significantly inferior performance of human judgment. The critical distinction in the experimental designs is that \citet{lawrence1985examination} let participants extrapolate the components of the time series, including trend and seasonality, through a tabular or graphical presentation, thus facilitating comparisons and the identification of seasonal patterns.

Noticing this discrepancy, \citet{edmundson1990decomposition} offers an application named GRAFFECT to structure judgmental forecasting. The authors test this application using a student sample and the 68 monthly time series from the M1 Competition. GRAFFECT assists users in initially identifying a linear trend with a potential of one change-point in history and an extrapolation slope defined by the user. Following this, users discern the perceived seasonal pattern from the de-trended data. This process can be repeated until users are satisfied that the residual contains nothing beyond noise. The forecasts produced through this approach significantly outperform unstructured judgmental forecasts. While they are also superior to any statistical method used in the M1 competition, the difference is not statistically significant.

Similarly, \citet{marmier2010structuring} develop a structured approach to guide forecasters in making adjustments to statistical forecasts based on expert knowledge of event effects. They identify four types of factors that can lead to significant and predictable changes in the forecast but are not accounted for by purely statistical methods: transient factors, transferred impact factors, quantum jump factors, and trend change factors. The authors test their approach in two case studies and find that the structured integration of managerial judgment leads to significant improvements in forecast accuracy compared to unaided adjustments and purely statistical forecasts.

\citet{trapero2013analysis} investigate the effectiveness of judgmental adjustments to statistical forecasts in the presence of promotional events.  The study finds that judgmental adjustments can improve forecast accuracy when dealing with promotions, particularly when the statistical model does not explicitly account for promotional effects. The effectiveness of these adjustments depends on various factors, such as the product type, promotional characteristics, and the underlying statistical model. The authors suggest that structured approaches to judgmental adjustments, which provide guidance on when and how to adjust based on the characteristics of the promotion and the statistical model, can lead to improved forecast accuracy.

The discussed research shows that judges perform considerably better when they handle a time series component-wise and that adjustment quality is contingent on the product time series and adjacent covariates.
We conclude that HNAM's composed forecasts promise firstly a higher acceptance due to a higher transparency and secondly more precise adjustments as they can happen on the level of individual covariate effects.

\subsection{Methodological Background}
\label{subsec:methodBackground}

This section introduces prior work that motivated HNAM.
That is, Meta's Prophet \citep{taylor2018forecasting} as a
highly interpretable yet arguably inflexible and narrowly applicable
model as well as Neural Additive Models (NAMs) \citet{agarwal2021neural}
for the idea of separately processing features with small neural networks before adding them in a linear model.
Finally, Neural Additive Time Series Models \citep{jo2023neural} follow a similar approach as ours with some important differences with respect to how sequences and feature interactions are handled. 

\paragraph{Prophet}
The Prophet algorithm, introduced by \citet{taylor2018forecasting} relies on a seasonal decomposition with covariates to create interpretable forecasts. Its accessible outputs and usability have generated considerable academic and practical interest as evidenced by 2455 citations according to Google Scholar and 17.7k stars on GitHub as per April 2024.

The Prophet forecast model is expressed as
$$y(t) = g(t) + s(t) + h(t) + e(t)$$
Here, each function corresponds to a unique aspect of the time series data:
$g(t)$ denotes the trend, encapsulating non-periodic changes in the time series. This is typically modelled as a piece-wise linear or logistic function.
$s(t)$ signifies seasonality, capturing periodic changes. This can include daily, weekly, or yearly cycles.
$h(t)$ accounts for the effects of holidays and additional covariates.
$e(t)$ represents the error term, capturing any idiosyncratic changes not modelled by the trend, seasonality, or holiday effects.

Prophet's ease of use and intuitive model structure makes it a popular choice for applied forecasting. At the same time, the simple structure can miss important non-linear effects and interactions. Further, Prophet has to be individually fitted for each time series, foregoing the benefits of cross-learning.

\paragraph{Neural Additive Models}
NAMs, as introduced by \citet{agarwal2021neural}, represent a class of models that harness the structure of generalized additive models with the ability to capture non-linear relationships characteristic of neural networks. 

The structure of the model can be denoted as:
$$f(x) = \sum_{i=1}^{d} f_i(x_i)$$
In this formulation, each term $f_i(x_i)$ represents an individual shallow neural network that is associated with the $i^{th}$ feature or input, $x_i$, on the response, $f(x)$. 

Accordingly, NAMs can capture non-linear effects. Additionally, their modularity facilitates the independent examination of each feature's effect on the outcome.
The additive structure inherent in NAMs provides a means to infer the effect sizes of each feature on the response variable, akin to the components $g(t)$, $s(t)$, and $h(t)$ in the Prophet model. 

\paragraph{Neural Additive Time Series Models}
\citet{jo2023neural} propose Neural Additive Time Series Models (NATMs). 
Similar to HNAM, NATMs adapt the concept of NAMs for time series forecasting.

NATMs differ from our approach in that they use a small neural net for every past timestep and for each feature. This makes NATMs potentially computationally expensive - an aspect the authors address by sharing weights across time or features.  NATMs, in their original formulation, are not modeling interactions across features and do not handle covariates known in the future.

\paragraph{Ex-post interpretability}
Research in interpretable and thereby more acceptable ML produced numerous tools to create ex-post explanations, such as LIME \citep{ribeiro2016should} or SHAP \citep{lundberg2017unified}. However, their  ex-post nature makes these methods computationally costly. At the same time, they still generate over-simplified approximations of the actual dynamics, leading scholars to call for inherently interpretable models \citep{rudin2019stop}. 

\subsection{Research Gap}
From the above, we conclude that present models are either
interpretable but oversimplifying or complex black-boxes that exacerbate algorithm aversion.
NAMs and NATMs provide promising steps towards interpretable yet sufficiently complex models. Our proposed HNAM architecture builds on these ideas
to yield interpretable forecasts that account for complex temporal dependencies and feature interactions.
\section{Proposed Architecture}
\label{sec:Architecture}
In describing the proposed model architecture of HNAM, we first present a conceptual overview and differentiate the roles of covariate types and the functional components processing them.
Afterwards, we describe our proposed implementation of the functional components in detail.

\subsection{Covariates and Components}
\begin{figure}[ht]
    \centering
    \includegraphics[width=1\linewidth]{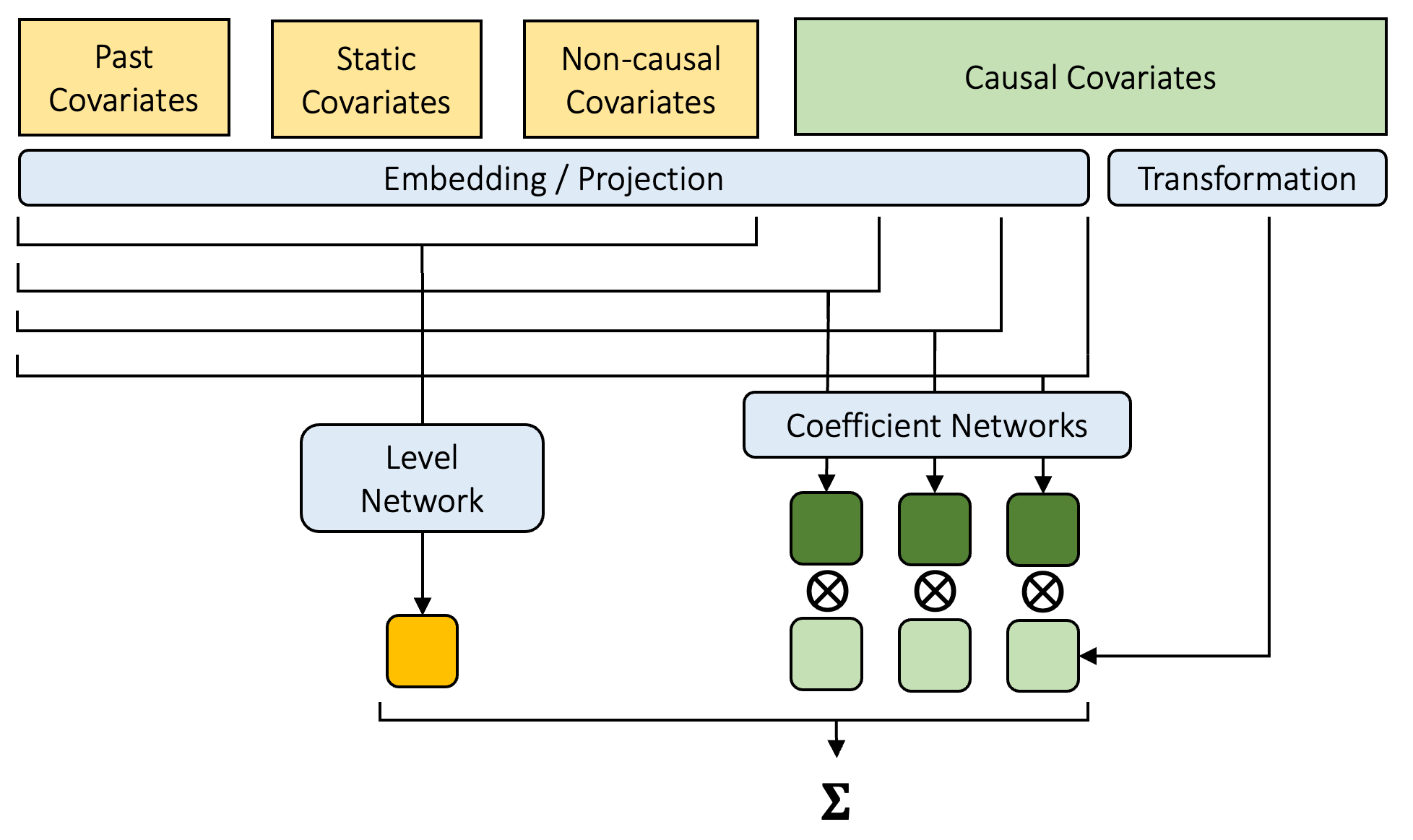}
    \caption{Conceptual Overview of the HNAM architecture}
    \label{fig:architecture}
\end{figure}

\paragraph{Conceptual overview} Figure \ref{fig:architecture} gives a high-level view on the architecture of HNAM. While all depicted concepts will be explained in later paragraphs, the general setup is as follows:
Covariates are differentiated with respect to their temporal availability and whether they should have an individual effect (i.e. be causal covariates).
They are embedded or projected (i.e. translated into a numerical representation of the model's internal dimensionality).
The \textit{level} network employs an Attention block \citep{vaswani2017attention} to project an estimated demand level into the forecasting horizon, using only information about the past.
One Attention-based \textit{coefficient} network per causal covariate processes the past, static, and non-causal covariates in addition to the respective causal covariate and those lower in the interaction hierarchy to derive one coefficient per timestep and causal covariate.
These coefficients are multiplied with the respective transformed values of the otherwise unprocessed covariates and all effects are added to the estimated level.

\paragraph{Causal covariates} HNAM is designed for demand forecasting situations, where some covariates that affect demand are known in the past and in advance. Examples of such covariates are the date (weekday, season, holiday) or planned promotional activities (price changes, advertisement). We call these \emph{causal covariates}. To address causal covariates, HNAM requires them to be specified as a hierarchical list, so that each covariate's estimated causal effect can depend on the covariate itself and those lower in the hierarchy. For the exemplary interaction hierarchy of \emph{Weekday, Promotion, Holiday}, the model estimates weekday effects independently of other causal covariates, promotional effects are estimated contingent on the weekday, and holiday effects are contingent on both weekday and promotional activity.
Accordingly, HNAM considers a matrix $C$ with one row per covariate, yielding $n_{c}$ rows ordered by the causal hierarchy, and $T_h + T_{f}$ columns for the concatenation of historical and future timesteps, respectively.
Note that our model does not prove or assume causality in the strict sense but differentiates covariates as to whether an analyst or manager deems that they \emph{should} have an individually estimated effect on demand.

\paragraph{Non-causal covariates} There are covariates that are known in the past and future that do not require individually estimated effects. Typical examples are the absolute and relative time indices and sinusoidal seasonal representations \citep[Chapter 12.1]{hyndman2021forecastingcomplex} that induce a temporal understanding to each timestep. This is necessary in non-recurrent neural networks \citep{vaswani2017attention}. Thereby, non-causal covariates modulate the effects of causal covariates whilst not having an individually estimated effect. Matrix $T$ stores these mostly time-related covariates across $T_h + T_{f}$ columns.

\paragraph{Static covariates} Similarly, matrix $S$ stores time-invariant covariates, typically product and store attributes that remain constant across $T_h + T_{f}$ columns.

\paragraph{Past covariates} Finally, matrix $P$ contains covariates that are only known in the past across $T_f$ columns. It typically contains the past demand observations alongside other covariates that are not known in advance.
Similar to non-causal and static covariates, past covariates have no individually estimated effects but also modulate causal covariate effects.
For more convenient indexing below, we let $P$ also have $T_h + T_{f}$ columns where values in the future are 0 and will not be used for forecasts.

\paragraph{Functional Components} In order to produce a forecast vector $y$ of $T_{f}$ forecast steps, the introduced covariate matrices pass through a \textit{level} neural network $g$ and \textit{coefficient} neural networks $f_{i}$ with one neural network for each of the $n_{c}$ causal covariates. Further, $t$ is a transformation function that k-1 one-hot encodes categorical causal covariates and standardizes continuous covariates. 

In order for the estimated level and coefficients to be meaningful, we must mask information accordingly.
The \textit{level} component only observes historical information and projects a future level for each time step in the forecasting horizon. In this way, it establishes the baseline demand level and captures possible trends.

Each coefficient neural network $f_i$ observes $S$, $T$, and $P$ completely, in addition to the $i$th causal covariate as well as causal covariates below it in the interaction hierarchy.
For continuous covariates, it outputs one coefficient; for categorical covariates it outputs $k-1$ coefficients where $k$ is their cardinality.

On a high-level, starting indexing at 0, HNAM can therefore be formulated as:
\begin{equation}
y = g(S,T,P) + \sum_{i = 0}^{n_{c}-1} f_{i}({S,T,P,C[:(i+1)]}) \cdot t({C[i,T_{h}:])}
\end{equation}

\subsection{Implementation}

In the following, we detail how functional components are implemented.
These encompass how covariates are processed via embeddings and projections, how covariate interactions are enabled by additively assembling information, and how temporal dependencies are modeled via Attention layers to determine the forecast's level and coefficients.

\paragraph{Embeddings and Projections} All covariates are projected into the model's embedding size, a hyperparameter that determines the model's internal dimensionality and thereby expressiveness.
To achieve this, an embedding table handles categorical covariates.
Embeddings are a sparser alternative to one-hot encodings to translate categorical values into a numerical representation.
Continuous covariates are projected into the model's embedding size through a linear-affine transformation.

\paragraph{Assembling information tensors} The constant embedding size of embedded covariates allows them to be summed in the feature dimension. Whenever HNAM requires a certain information set, the respective covariates are summed and then passed through a layer normalization.

\paragraph{MLP} At multiple points in our architecture, a multi-layer perceptron is responsible for non-linear processing.
The input is projected into an embedding size that is larger by a user-parameterized factor and then passes through a Gaussian Error Linear Unit (GELU) activation function for non-linearity. Subsequently, the representation is projected back with dropout to the network's embedding size.

\paragraph{Temporal Convolution Layer} This layer expands its input using a 1-dimensional convolution with a kernel size of 3, accounting for the last two timesteps with zero padding for the two oldest observations. Post-convolution, a GELU activation and dropout regularize the output before it is projected back to the embedding size.
This layer is used in Attention blocks described below to extract local features across adjacent timesteps.

\paragraph{Attention Block} Both \textit{level} and \textit{coefficient} networks incorporate information from other time steps globally via multi-head Attention \citep{vaswani2017attention} with a temporal convolution to calculate queries, keys, and values. The respective implementations are described below.

\paragraph {Level Network} The \textit{level} component is implemented as an Attention block. The query consists of of $(S[:,T_h:]+T[:,T_h:])$, i.e., static and temporal information for the forecast horizon. The keys and values both are $(S+T+P)[:T_h]$, i.e., static, temporal, and past information in the past time steps.

\begin{itemize}
    \item Query ($Q$), Key ($K$), and Value ($V$) projections are obtained by passing the input sequence through temporal convolutional layers: 
    \begin{align*}
    Q &= \text{TemporalConv}((S+T)[:,T_h:]), \\
    K &= \text{TemporalConv}((S+T+P)[:,:T_h]), \\
    V &= \text{TemporalConv}((S+T+P)[:,:T_h])
    \end{align*}

    \item The sequence is split in the channel dimension into multiple heads and scaled dot-product Attention is computed for each head:
    \begin{align*}
    Y &= \text{Attention}(Q, K, V) \\
      &= \text{Softmax}\left(\frac{QK^T}{\sqrt{d_k}}\right)V
    \end{align*}
    where $d_k$ is the dimension of each head.
    
    \item The attention output is then combined with the output of an MLP after layer normalization:
    \begin{align*}
    Y' &= Y + \text{MLP}(\text{LayerNorm}(Y))
    \end{align*}
    
    \item Finally, the Level vector and a Level Embedding in the embedding size are projected from $Y'$:
    \begin{align*}
    \text{Level} &= \text{Linear}(Y') \\
    \text{Level Emb} &= \text{Linear}(Y')
    \end{align*}
\end{itemize}

\paragraph{Coefficient Networks}
For each causal covariate, we have one \textit{coefficient} network. A similar Attention block as described above is at the core of each \textit{coefficient} network.
Queries, keys and values for the $i$th covariate and \textit{coefficient} network are obtained as:
    \begin{align*}
    Q &= \text{TemporalConv}((S+T+C[:i+1])[T_h:] + \text{Level Emb}), \\
    K &= \text{TemporalConv}((S+T+P+C[:i+1])[:T_h]), \\
    V &= \text{TemporalConv}((S+T+P+C[:i+1])[:T_h]))
    \end{align*}
As described previously, for each causal covariate's \textit{coefficient} network, the values of causal covariates higher in the interaction hierarchy are not available. We add the Level Embedding obtained earlier to the query to utilize the extracted information on the level here.

Although each feature is processed in an individual network, we let Attention scores be computed in parallel by assembling tensors such that the different causal covariates are stacked in a second batch dimension.
The results of the Attention block are projected back to obtain the coefficients.
\section{Datasets and Preprocessing}
\label{sec:datasets}

For robust benchmarking, we generate and evaluate forecasts in three different retail demand datasets.
All datasets have multiple covariates alongside the recorded sales data.
Two datasets are publicly available and one stems from an industry partner.

In the following, we provide an overview of the preprocessing we apply in this study, highlight how time series are filtered, what covariate features are used, and how a rolling-forward evaluation is carried out.

\subsection{Datasets}

\paragraph{Walmart}
This public dataset \citep{m5comp} consists of hierarchical sales data provided by Walmart through the University of Nicosia amid the M5 forecasting competition. It spans 1941 days from 2011 to 2016. The dataset includes daily item-level sales data of 3049 products and 10 store locations across three states.

\paragraph{Favorita}
The Favorita dataset \citep{favorita2018} is a publicly available dataset provided by Corporación Favorita, an Ecuadorian-based grocery retailer, for a forecasting competition. It spans 1684 days from 2013 to 2017 across 4035 products and 54 stores.

\paragraph{Retail}
We further evaluate models on a dataset provided by a medium-sized retailer. It spans 2472 days from 2015 to 2022 observing 2910 products across three stores.

\subsection{Data Selection}
The research documented in this paper focuses on the interpretability aspect of forecasting to enhance human-computer interaction. To create a fitting benchmarking environment for this purpose, we select for time series, i.e. a unique product and store combination, that are likely to be adjusted due to their economic relevance and for which we can assess the plausibility of interpretability outputs. 

We exclude time series with apparently low, intermittent, or cyclical sales.
All datasets lack pertinent information as to whether a product was out of stock or not offered at all, making it challenging to create realistic benchmarks for forecasting consumer demand. Since we can not differentiate these scenarios without making substantial assumptions, we focus on products with more stable sales, assuming they were less frequently out of stock.

Further, we argue that daily forecasts of intermittently demanded products are generally less relevant than non-intermittent ones. For instance, products with high daily demand and short shelf-lives, such as fruits, vegetables, and dairy products, typically do not exhibit intermittency. On the other hand, products with slower turnover rates, often due to higher costs or specificity, usually have longer shelf-lives, like spices, beverages, and non-food items. For these products, aggregate forecasts are more relevant, making the problem of intermittency conveniently disappear.

Based on these arguments, we select time series from the given data sets according to the following criteria:
\begin{enumerate}
    \item The time series must display non-zero sales for at least 100 days prior to the first test period.
    \item The time series must not exhibit more than 100 days of no sales after the first recorded sale.
    \item The time series' median sales must be greater than five during the test periods and 100 days before.
    \item The time series must be in the top 500 in terms of sales during the test periods and 100 days before.
    \item The time series must be in the top 500 in terms of revenue during the test periods and 100 days before (not applicable for Favorita since price information is missing).
    \item For Favorita, we instead select time series with less than 1\% missing observations.
\end{enumerate}

This selection process yields 221 (Walmart), 287 (Favorita), and 193 (Retail) time series.

\subsection{Features}

The multivariate models we benchmark utilize a number of covariates. While most covariates are available in all datasets, some are dataset specific. Unless indicated otherwise, all covariates are available not only for past observations but also in the prediction horizon. Table \ref{tab:features} lists the covariates, differentiates their type as categorical or continuous and indicates their availability per dataset. In addition to the listed covariates, neural networks use an absolute and a relative time index as well as sine and cosine embeddings of the day of the year to induce a temporal sense to the models.

\begin{table}[h!]
\centering
\begin{threeparttable}
\caption{Covariates entering multivariate models per dataset.}
\label{tab:features}
\begin{tabular}{@{}lccc@{}}
\toprule
 & M5 Walmart & Favorita & Retail \\
\midrule
\multicolumn{4}{@{}l@{}}{Categorical} \\
\cmidrule(r){1-4}
Product & \cmark & \cmark & \cmark \\
Store & \cmark & \cmark & \cmark \\
Promotion & \xmark & \cmark & \cmark \\
Weekday & \cmark & \cmark & \cmark \\
Holiday\tnote{1} & \cmark & \cmark & \cmark \\
SNAP\tnote{2} & \cmark & \xmark & \xmark \\
\midrule
\multicolumn{4}{@{}l@{}}{Continuous} \\
\cmidrule(r){1-4}
Sales (only past) & \cmark & \cmark & \cmark \\
Relative Price \tnote{3} & \cmark & \xmark & \cmark \\
Oil Price (only past) & \xmark & \cmark & \xmark \\
\bottomrule
\end{tabular}
\begin{tablenotes} \footnotesize
\item[1] For Favorita, we differentiate local, regional, and national holidays as separate covariates; for Walmart we differentiate sporting, cultural, and national events or holidays.
\item[2] The Supplemental Nutrition Assistance Program (SNAP), previously known as the Food Stamp Program, provides funds specifically for purchasing food to low-income households. The feature marks a day at which funds are distributed.
\item[3] The percentage deviation of the day's price from the rolling 20-day mean.
\end{tablenotes}
\end{threeparttable}
\end{table}

\subsection{Evaluation}
Across all datasets, we use the last five fully available months as test sets.
For each test month, we train neural networks with observations up to two weeks prior to the test month and use the remaining two weeks prior as validation set for early stopping.

We generate forecasts with a maximum horizon of two weeks via a sliding window in each test month without retraining.
We fine tune with updated data prior to each new test month.
Similarly, we keep the hyperparameters for the remaining benchmark models constant per test month but use all data up to the respective forecast date for training.

We aim to forecast sales with daily granularity. While coarser temporal aggregations can be sensible depending on the business case, we focus on our model's interpretable forecasts based on covariates of daily frequency.
Further, we limit our exploration to forecasting expected sales, as opposed to providing probabilistic forecasts as required for newsvendor models. While HNAM supports quantile prediction, we opt for point forecasts for the sake of clearer error metrics and interpretability outputs in this paper.
\section{Models}
\label{sec:models}

The following provides an overview of the considered benchmarks and implementation details. To that end, we first describe variants of HNAM implemented for the various target sets, before, detailing our implementation of Temporal Fusion Transformers, Prophet, ETS, and SARIMAX. 

\paragraph{Hierarchical Neural Additive Models}
HNAM's architecture has been outlined in Section \ref{sec:Architecture}.
There, we discriminated different types of covariates and introduced the interaction hierarchy applied for the benchmark data sets.

We use the same hierarchy for all datasets, given that the respective covariate is available:
\emph{Weekday, Relative Price, Promotion, Holiday}.
We select this hierarchy because it feels natural to estimate weekday effects
independently of other modulating covariates and add more sparsely occurring covariates on top.
Generally speaking, we suggest to order covariates by decreasing frequency of their occurrence to limit the total amount of covariate interactions.

The Walmart data set does not include information on scheduled promotions.
However, \emph{SNAP} indicates days where eligible customers received financial support for groceries.
Holidays are differentiated into three separate covariates in Walmart and Favorita, differentiating their type and regionality respectively.

\paragraph{Temporal Fusion Transformers}
Temporal Fusion Transformers (TFT) \citep{lim2021temporal} is a deep learning model that employs the transformer architecture for time series forecasting. Its introductory paper demonstrated TFT outperforming other popular neural networks for forecasting, including DeepAR \citep{salinas2020deepar}.
Like HNAM, TFT handles and differentiates covariates with respect to their availability during the forecast horizon. Notably, its attention mechanism and variable selection networks offer some interpretability that is, however, significantly less immediate than in our proposed architecture \citep{feddersen2023solving}.

\paragraph{Neural Network Implementations} The neural network-based methods above are implemented in Pytorch Forecasting \citep{githubGitHubJdb78pytorchforecasting}.
In pre-experiments, we observed better validation scores with an embedding size of 32 rather than 16 in both HNAM and TFT. 
Otherwise, we use the package's default implementations and parameters for TFT.
We do not perform automatic hyperparameter tuning due to its exponential computational costs and because predictive accuracy is not the main proposition of our proposed architecture.
Rather, our aim is to validate whether constraints imposed on HNAM for interpretability would reduce its predictive accuracy notably compared to more opaque models like TFT.

Neural network models learn global models for all time series in a dataset whereas the benchmark methods below fit local models per time series. Global models can thereby offset their complexity (number of parameters) with a relatively larger amount of training data (fitting only one model per dataset instead of one model per time series). 
Accordingly, we fit the neural network models with a learning rate of 0.001, weight decay of 0.01 and a batch size of 256.
For the first test month, we train up to 300 epochs with early stopping on the loss in the validation set (i.e., the last two weeks of each training set) with a patience of 30 epochs and select the best model with respect to the validation loss.
For the remaining test months, neural networks are fine-tuned for up to 100 epochs, also employing early stopping.
The models are trained on a high-performance cluster node with an NVIDIA V100 GPU.

\paragraph{Prophet}
Prophet \citep{taylor2018forecasting} decomposes a time series into trend, seasonality, and holiday components. The model conveniently handles missing values and is intended to allow fast human-in-the-loop iterations. We use Prophet's Python implementation \citep{facebookProphet} with standard parameters. Per time series we validate an additive versus a multiplicative model in the training data and use the better performing model for generating forecasts.

\paragraph{ETS}
Seasonal exponential smoothing (also termed ETS for Error, Trend, Seasonality) applies exponentially decaying weights to the error, trend, and seasonal components of univariate time series.
We use the statsforecast \citep{garza2022statsforecast} AutoETS implementation to automatically determine the additive or multiplicative configuration of error, trend, and seasonality via the Akaike Information Criterion.

\paragraph{SARIMAX}
Seasonal Autoregressive Integrated Moving-Average with Exogenous Regressors (SARIMAX) models understand the detrended time series as a combination of autoregressive and moving average processes, alongside a seasonal component and external regressors. The respective parameters are found with the statsforecast AutoARIMA implementation.
\section{Results}
\label{sec:results}
\subsection{Computation Times}

Table \ref{tab:training_details} depicts details on the training process of the neural network models HNAM and TFT for the first test set of each data set.
Although both models run for similar numbers of epochs until early stopping triggers
and achieve comparable loss scores in the training and validation sets, HNAM requires considerably less time per epoch. Accordingly, it needs markedly shorter overall training times than TFT.

As to be expected, the linear models train notably faster, with ETS taking up to ten minutes, Prophet up to two hours, and SARIMAX up to four hours on an M1 Macbook Pro to generate all forecasts.
One should note, however, that a pre-trained neural network requires less training time for ongoing fine-tuning and that inference is possible in a few minutes.

\begin{table}[ht]
\centering
\caption{Training times and loss metrics for HNAM and TFT.}
\label{tab:training_details}
\resizebox{\linewidth}{!}{
\begin{tabular}{@{}lcccccc@{}}
\toprule
& \multicolumn{2}{c}{Walmart} & \multicolumn{2}{c}{Favorita} & \multicolumn{2}{c}{Retail} \\
\cmidrule(r){2-3} \cmidrule(lr){4-5} \cmidrule(l){6-7}
& HNAM & TFT & HNAM & TFT & HNAM & TFT \\
\midrule
Time per epoch (m:s) & 01:57 & 03:44 & 01:30 & 03:12 & 01:32 & 05:02 \\
Epochs & 258 & 253 & 220 & 261 & 211 & 217 \\
Training time (h:m:s)& 8:23:06 & 15:44:32 & 5:30:00 & 13:55:12 & 5:23:32 & 18:12:14 \\
Training loss & 0.650 & 0.636 & 0.537 & 0.517 & 0.572 & 0.569 \\
Validation loss & 0.601 & 0.669 & 0.518 & 0.500 & 0.543 & 0.550 \\
\bottomrule
\end{tabular}
}
\end{table}

\subsection{Aggregate Error Metrics}

This section presents aggregate error metrics of the considered models in the three dataset's test sets.
Negative predictions are truncated at 0.
In terms of error metrics, we consider SMAPE, MAE and RMSE. We select SMAPE for its intuitive interpretation noting its asymmetric weighing of over- and underprediction. MAE as standardized with each time series' standard deviation provides a common linear and symmetric metric. We chose the equally standardized RMSE as a loss metric in the training process. Finally, we also consider the frequency of 1\textsuperscript{st} and 2\textsuperscript{nd} ranks achieved across time series
when ranking models by RMSE per time series.

\begin{table}[ht]
\centering
\caption{Mean and median error metrics in Walmart}
\label{tab:metrics_median_walmart}
\resizebox{\linewidth}{!}{
\begin{tabular}{rcccccccc}
\toprule
{} & \multicolumn{2}{l}{SMAPE} & \multicolumn{2}{l}{Std. MAE} & \multicolumn{2}{l}{Std. RMSE} & \multicolumn{2}{l}{Rank Freq.} \\
{} & $\bar{x}$ & $\tilde{x}$ & $\bar{x}$ & $\tilde{x}$ & $\bar{x}$ & $\tilde{x}$ & 1\textsuperscript{st} & 2\textsuperscript{nd} \\
\midrule
HNAM    &     0.297 &       0.211 &     0.571 &       0.451 &     4.000 &       1.387 &                 76.5\% &                 23.5\% \\
TFT     &     0.310 &       0.222 &     0.601 &       0.485 &     4.304 &       1.611 &                 23.5\% &                 75.6\% \\
PROPHET &     0.412 &       0.307 &     0.838 &       0.653 &     9.590 &       2.897 &                  0.0\% &                  0.5\% \\
SARIMAX &     0.400 &       0.296 &     0.796 &       0.635 &     8.045 &       2.747 &                  0.0\% &                  0.5\% \\
ETS     &     0.373 &       0.273 &     0.748 &       0.594 &     7.190 &       2.384 &                  0.0\% &                  0.0\% \\
\bottomrule
\end{tabular}
}
\end{table}

\begin{table}[ht]
\centering
\caption{Mean and median error metrics in Retail}
\label{tab:metrics_median_retail}
\resizebox{\linewidth}{!}{
\begin{tabular}{rcccccccc}
\toprule
{} & \multicolumn{2}{l}{SMAPE} & \multicolumn{2}{l}{Std. MAE} & \multicolumn{2}{l}{Std. RMSE} & \multicolumn{2}{l}{Rank Freq.} \\
{} & $\bar{x}$ & $\tilde{x}$ & $\bar{x}$ & $\tilde{x}$ & $\bar{x}$ & $\tilde{x}$ & 1\textsuperscript{st} & 2\textsuperscript{nd} \\
\midrule
HNAM    &     0.414 &       0.294 &     0.428 &       0.311 &     8.892 &       1.597 &                 47.2\% &                 50.3\% \\
TFT     &     0.407 &       0.293 &     0.427 &       0.315 &     8.294 &       1.637 &                 52.3\% &                 46.6\% \\
PROPHET &     0.555 &       0.438 &     0.674 &       0.466 &    22.782 &       3.551 &                  0.0\% &                  0.5\% \\
SARIMAX &     0.587 &       0.465 &     0.715 &       0.493 &    25.649 &       3.983 &                  0.0\% &                  1.0\% \\
ETS     &     0.541 &       0.416 &     0.647 &       0.434 &    23.738 &       3.060 &                  0.5\% &                  1.6\% \\
\bottomrule
\end{tabular}
}
\end{table}

\begin{table}[ht]
\centering
\caption{Mean and median error metrics in Favorita}
\label{tab:metrics_median_favorita}
\resizebox{\linewidth}{!}{
\begin{tabular}{rcccccccc}
\toprule
{} & \multicolumn{2}{l}{SMAPE} & \multicolumn{2}{l}{Std. MAE} & \multicolumn{2}{l}{Std. RMSE} & \multicolumn{2}{l}{Rank Freq.} \\
{} & $\bar{x}$ & $\tilde{x}$ & $\bar{x}$ & $\tilde{x}$ & $\bar{x}$ & $\tilde{x}$ & 1\textsuperscript{st} & 2\textsuperscript{nd} \\
\midrule
HNAM    &     0.202 &       0.139 &     0.430 &       0.319 &    20.580 &       5.205 &                 29.6\% &                 68.6\% \\
TFT     &     0.191 &       0.135 &     0.413 &       0.312 &    17.382 &       4.942 &                 70.0\% &                 30.0\% \\
PROPHET &     0.288 &       0.198 &     0.643 &       0.474 &    98.417 &      11.485 &                  0.0\% &                  0.0\% \\
SARIMAX &     0.282 &       0.204 &     0.637 &       0.471 &    49.997 &      11.361 &                  0.0\% &                  0.3\% \\
ETS     &     0.261 &       0.185 &     0.591 &       0.434 &    45.212 &       9.617 &                  0.3\% &                  1.0\% \\
\bottomrule
\end{tabular}
}
\end{table}

Overall, we note that the neural network models HNAM and TFT clearly outperform the linear benchmarks.
Comparing the two, HNAM performs better in the Walmart dataset (Table \ref{tab:metrics_median_walmart}), whereas TFT dominates in the Favorita dataset (Table \ref{tab:metrics_median_favorita}),
while Retail shows balanced outcomes slightly favoring TFT (Table \ref{tab:metrics_median_retail}).

These results confirm that the considerably less flexible HNAM architecture is competitive
with a state-of-the-art model that employs more complex components and takes considerably longer to train.
As an additional interesting finding, SARIMAX and Prophet, despite utilizing covariates, fare worse than the univariate ETS,
once again confirming ETS as a hard-to-beat benchmark.

\subsection{Interpretability}

After confirming HNAM's competitive accuracy on a macrolevel, this section explores specific forecasting instances. This examination will highlight HNAM's ability to offer explanations for its forecasts, as well contrasting its accuracy versus TFT in favorable and infavorable instances.

\begin{figure}[htbp]
    \centering
    \begin{subfigure}{1\linewidth}
        \includegraphics[width=\linewidth]{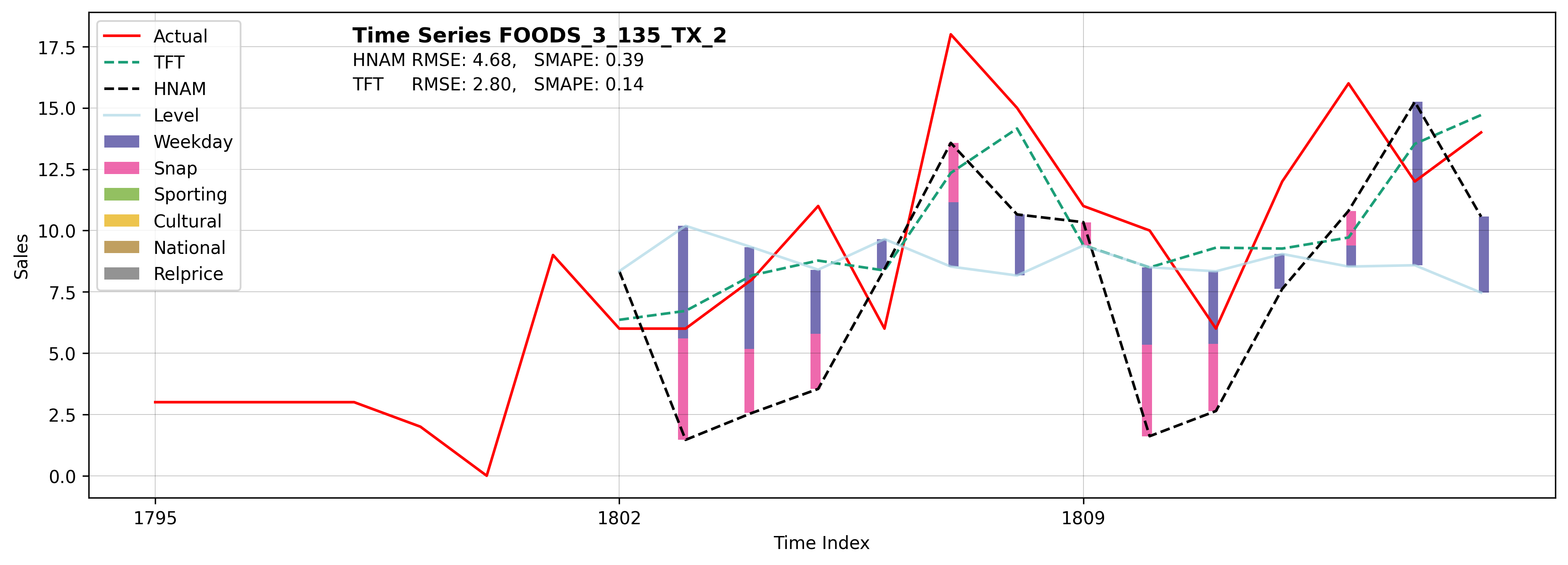}
        \caption{Relatively inaccurate HNAM predictions for one forecast instance in Walmart}
        \label{fig:walmart-1}
    \end{subfigure}
    \hfill 
    \begin{subfigure}{1\linewidth}
        \includegraphics[width=\linewidth]{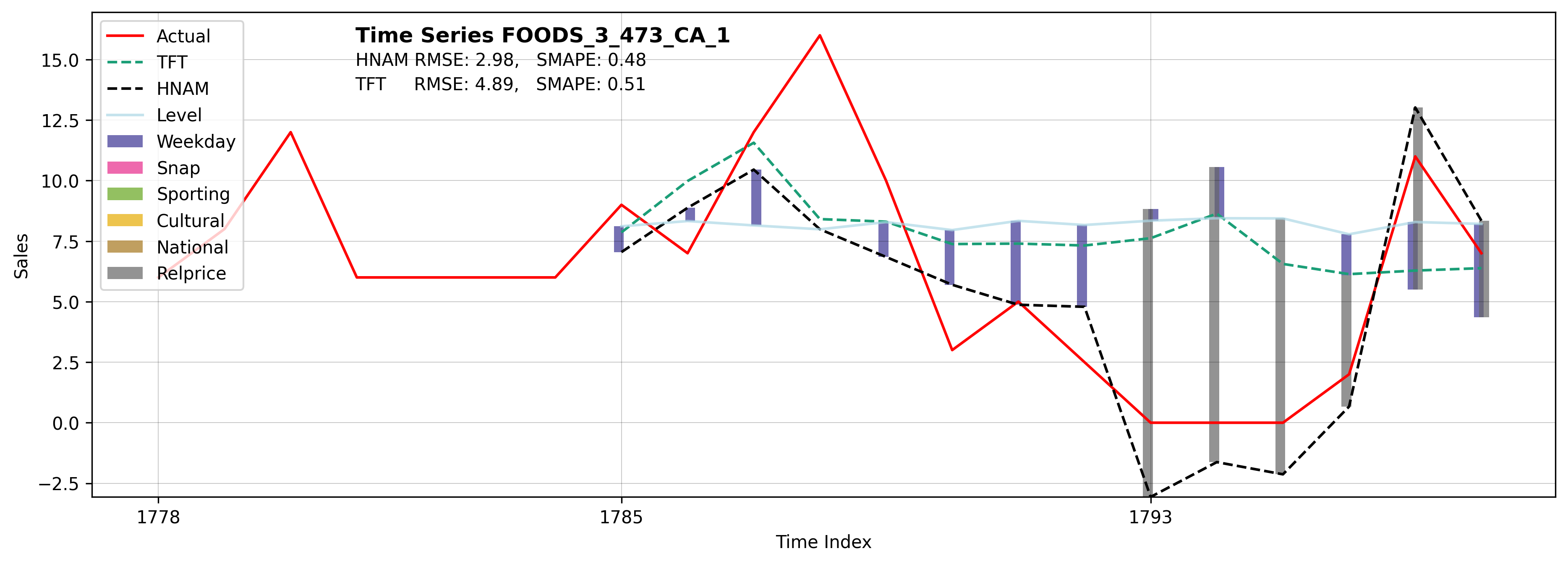}
        \caption{Relatively accurate HNAM predictions for one forecast instance in Walmart}
        \label{fig:walmart-2}
    \end{subfigure}
    \caption{Composed HNAM predictions and TFT predictions versus actuals in Walmart.}
    \label{fig:walmart-forecasts}
\end{figure}

Figure \ref{fig:walmart-1} visualises a forecast instance from the Walmart dataset, where HNAM fares considerably worse than TFT as measured by the largest gap between the two in SMAPE. Timestep 1802 on the the x-axis marks the first date of the forecast horizon. The y-axis measures sales quantities.
To explain an HNAM forecast, one first considers the predicted Level (light-blue line). Causal covariate effects are then added to that Level (colored stacked bars) and amount to the model's forecast (dashed black line).
In the depicted instance, the first forecast at timestep 1802 is determined solely by the Level, devoid of any causal covariate effects. This occurs due to the application of k-1 dummy encoding for categorical variables, where the first category is represented implicitly by the absence of other categories (all zeros in the dummy-encoded vector).
We state all potentially considered covariates in the legend, even if they have no effect in the given forecast horizon. Such a representation clarifies that the covariates would, in fact, be considered if they had a non-zero effect. The TFT prediction is given as a dashed blue line for comparison.

Comparatively bad performances of HNAM as observed in Figure \ref{fig:walmart-1} often occur after rapid level changes, however, we do not have a concise explanation at hand that goes beyond suspecting that HNAMs composed structure makes it more inflexible for picking up such rapid shifts.

Figure \ref{fig:walmart-2} depicts an instance where HNAM markedly outperforms TFT.
Particularly, the initially demand-reducing effect of a substantial price increase as well as a subsequent rebound following a slight price decrease is accurately captured by HNAM. We observe how the additive nature of HNAM's predictions arguably make it more prone for producing negative predictions which would have to be truncated.

\begin{figure}[htbp]
    \centering
    \includegraphics[width=\linewidth]{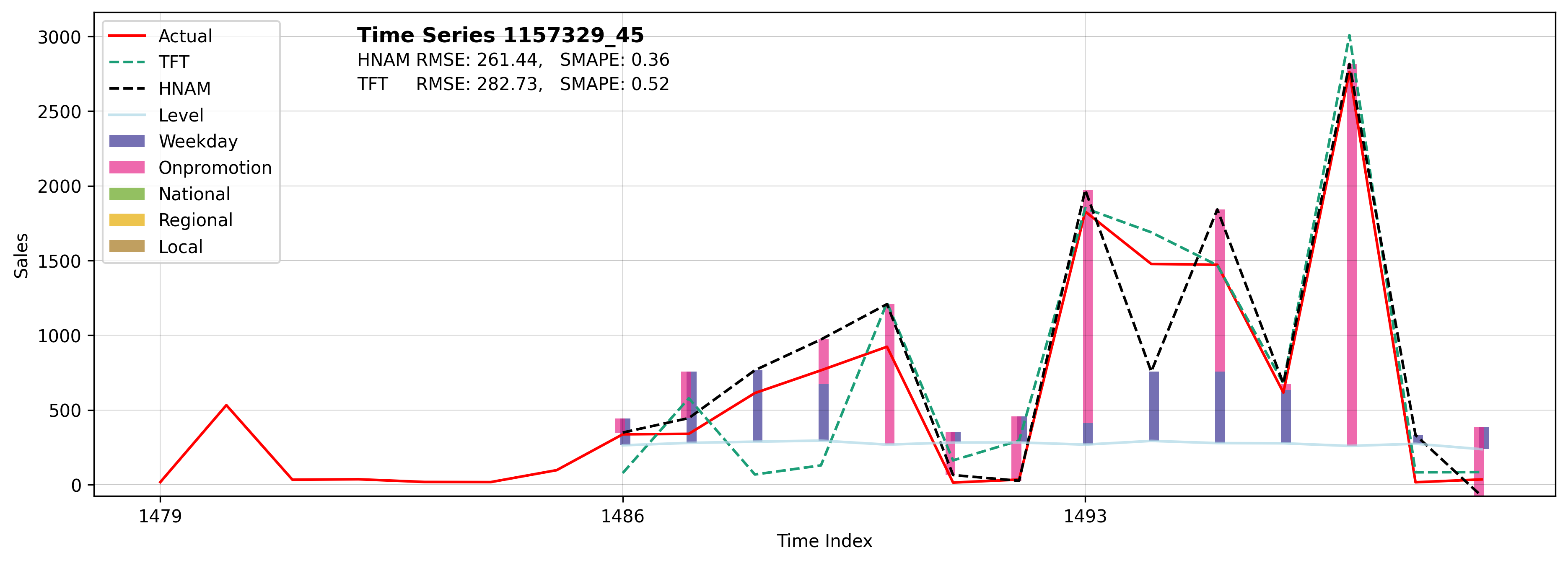}
    \caption{Composed HNAM predictions and TFT predictions versus actuals in Favorita.}
    \label{fig:favorita-1}
\end{figure}

Figure \ref{fig:favorita-1} depicts an instance where HNAM performs relatively well and apparently
predicts the promotional effects quite well. This instance is an example of its strength and weakness regarding the clear attribution to covariates. Considering timestep 1494, the data does not indicate a planned promotion for that day although it is surrounded by days with active promotions.
Yet, the actual sales remain relatively high which is accurately predicted by TFT.
Assuming this is a regular, learnable pattern, HNAM would have no chance to do so since
the estimated coefficient for promotional activity for that day is multiplied with zero anyway.

\section{Conclusions and Future Research}
\label{sec:conclusions}

\paragraph{Accuracy results} This study introduces Hierarchical Neural Additive Models (HNAM) for time series forecasting emphasising the critical aspect of interpretability.
The accuracy results observed in the computational study demonstrate that HNAM delivers competitive forecasting accuracy when benchmarked against advanced models such as Temporal Fusion Transformers (TFT) across multiple datasets. This is particularly noteworthy as HNAM incorporates design elements aimed at enhancing interpretability, which could potentially restrict the model's expressive power.
In addition, HNAMS's composed forecasts allow for precise attributions how different covariates impact the forecast outcomes.
This interpretability can be used for designing user interfaces that foster trust and encourage the productive
injection of extra-model knowledge by managers.

\paragraph{Possible challenges in practice} Although HNAM marks a big step towards inherently interpretable models for demand forecasting, we anticipate common problems upon its practical implementation for decision-support:

While HNAM's architecture allows a clear mapping of which information the model used to estimate a covariate effect,
one must not confuse this with proving causality. At least two examples come to mind:
In the Retail dataset, we have two covariates regarding promotional activity: promotion and price change.
Some products have promotions without price changes and vice versa. Given a product that only has price changes
concurrent with promotions, the estimated price sensitivity might counterintuitively show a negative effect
of the price change if it is of a lesser extent than other price changes during promotions.
One solution would be to drop the promotion covariate entirely or to use it only when there are no price changes.
As a similar example, it is conceivable that past price increases were a reaction to anticipated increased demand.
Then, the price sensitivity would again imply higher demand for higher prices. Hence, the usual
caveats of statistical modeling apply to HNAM and estimated effects must be carefully evaluated before
concluding any causality to rest pricing decisions upon.

\paragraph{Behavioral validation} 
The present research rests upon the arguments laid forth in Section \ref{sec:Motivation} to motivate
HNAM's potential for improving human-computer interaction in the retail demand prediction context.
However, no study to date has clearly examined the case where analysts work with a black-box versus a composed forecasting model. Therefore, we plan to conduct and encourage future behavioral research on analysts' interactions with explainable models like HNAM.
These would optimally include discussions and quantitative evaluations with domain-experts as well as larger-scale evaluations with business students.

\paragraph{Architecture} Our HNAM implementation uses several best practices (e.g., layer normalization, dropout, multi-head Attention), yet it is by no means the result of an extensive architecture search.
We emphasise that the general idea of HNAM might be implemented in more efficient and more accurate ways.

\paragraph{Delayed and anticipatory effects} 
As alluded to earlier when discussing Figure \ref{fig:favorita-1}, HNAM in its current implementation does not natively model delayed or anticipatory effects. That is, for example, when a promotion affects subsequent
periods \citep{hewage2022forecast}, or when customers change their shopping patterns in anticipation of an event or holiday. Such effects, if suspected, could be captured by adding suitably engineered covariates with
positive and negative shifts. Future research, however, might further consider neural architectural solutions that maintain the clear effect attribution of HNAM while inherently modeling such time-shifted covariate effects.

\paragraph{Final Remarks}
As organizations contend with data-rich environments and challenges with algorithm aversion,
our results demonstrate that HNAM makes a significant step towards more effective decision-making through accurate yet explainable models.
We call for behavioral research that refines the concept of algorithm aversion as to how it may be attenuated
through inherently interpretable models like the one presented here.

\bibliography{bib}
\bibliographystyle{apalike}
\end{document}